\documentclass{article}
\usepackage{arxiv}

\usepackage[usenames,dvipsnames,svgnames,table]{xcolor}
\usepackage{xspace}

\usepackage{url}  
\usepackage{graphicx}  
\usepackage{amsmath,amssymb,amsthm}
\usepackage{amsfonts}

\usepackage{enumitem} 

\usepackage{cite}

\usepackage [english]{babel} 
\usepackage [autostyle, english = american]{csquotes}
\MakeOuterQuote{"}

\usepackage[normalem]{ulem} 
\useunder{\uline}{\ul}{}    

\usepackage{microtype}
\usepackage{graphicx}
\usepackage{subcaption} %
\usepackage{booktabs} 

\usepackage{algorithm}
\usepackage{algpseudocode}
\algrenewcommand\algorithmicrequire{\textbf{Input:}}
\algrenewcommand\algorithmicensure{\textbf{Output:}}

\usepackage[pagebackref=true,breaklinks=true,letterpaper=true,colorlinks,bookmarks=false]{hyperref}

\newtheorem{definition}{Definition} 

\graphicspath{{./}{./figures/}}
\DeclareGraphicsExtensions{.pdf,.jpeg,.png,.jpg,.eps}

\def\onedot{\ifx\@let@token.\else.\null\fi\xspace}
\def\eg{\emph{e.g}\onedot} 
\def\eg{\emph{e.g}\onedot} 
\def\ie{\emph{i.e}\onedot}

\def\wrt{w.r.t\onedot} 
\def\etal{\emph{et al}\onedot}
\makeatother

\def\Vec#1{{\boldsymbol{#1}}} 
\def\Mat#1{{\boldsymbol{#1}}} 
 
\def\GRASS#1#2{\mathcal{G}({#2},{#1})} 
\def\ST#1#2{\mathrm{St}({#2},{#1})} 
\def\ORTHO#1{\mathcal{O}({#1})} 

\DeclareMathOperator{\Tr}{Tr}

\title{A Probabilistic approach for Learning Embeddings without Supervision\thanks{Preprint.}}

\author{Ujjal Kr Dutta,\textsuperscript{$\dag$,1} Mehrtash Harandi,\textsuperscript{*} C Chandra Sekhar\textsuperscript{$\dag$}\\
\textsuperscript{$\dag$}Dept. of Computer Science and Eng., Indian Institute of Technology Madras, India\\
\textsuperscript{*}Dept. of Electrical and Computer Systems Eng., Monash University, Australia\\
{\small \textsuperscript{1}ukd@cse.iitm.ac.in}
}

\begin{document}
\maketitle

\begin{abstract}
For challenging machine learning problems such as zero-shot learning and fine-grained categorization, embedding learning is the machinery of choice because of its ability to learn generic notions of similarity, as opposed to class-specific concepts in standard classification models. Embedding learning aims at learning discriminative representations of data such that similar examples are pulled closer, while pushing away dissimilar ones. Despite their exemplary performances, supervised embedding learning approaches require huge number of annotations for training. This restricts their applicability for large datasets in new applications where obtaining labels require extensive manual efforts and domain knowledge. In this paper, we propose to learn an embedding in a completely unsupervised manner without using any class labels. Using a graph-based clustering approach to obtain pseudo-labels, we form triplet-based constraints following a metric learning paradigm. Our novel embedding learning approach uses a probabilistic notion, that intuitively minimizes the chances of each triplet violating a geometric constraint. 
Due to nature of the search space, we learn the parameters of our approach using Riemannian geometry. Our proposed approach performs competitive to state-of-the-art approaches.
\end{abstract}


\maketitle

\section{Introduction}
\label{sec:introduction}

Embedding learning is the machinery of choice in many challenging machine learning problems where standard classification models cannot be used with ease. For example, softmax-based classification networks using cross-entropy loss are hard to train when the number of classes is huge (\textit{extreme classification} \cite{yen2016pd,prabhu2014fastxml}). For cross-domain tasks \cite{Chen_2019_CVPR,You_2019_CVPR,yu2016sketch}, classification-based models require complicated training procedures. 
Also, for the challenging \textit{zero-shot learning scenario} \cite{ZSL_good_bad_ugly, schonfeld2019generalized} where the test examples belong to semantic classes not seen during training, embedding learning is preferred over standard classification models. This is because of its ability to capture \textit{generic notions} of similarity, rather than \textit{class-specific concepts}. Furthermore, classification-based models cannot handle probabilistic class labels, but embedding learning can \cite{huai2018metric}.

Learning rich and discriminative representations of data is essential in various problems, including the ones mentioned above. Embedding learning aims at learning representations or \textit{embeddings} of data, while grouping \textit{similar} examples and segregating \textit{dissimilar} ones. As the common practice is to learn a metric in the embedding space, \textit{embedding learning} is often studied interchangeably with \textit{metric learning} \cite{HDML_CVPR19, Deep_AML, Bellet2015}.
Metric learning methods require only a \textit{weaker form of supervision}, usually provided as \textit{pairs} \cite{ITML, KISSME, GMML, LR-GMML, AML, MDMLCLDD}, \textit{triplets} \cite{LMNN, FaceNet, SCML, DRIFT}, \textit{batches} \cite{Lifted_structure} or \textit{tuples} \cite{N_pair}. This is yet another advantage of metric learning over classification-based models.

However, in practice, pairs or triplets are constructed from class labels, thus making commonly used metric learning approaches \cite{LR-GMML, AML, MDMLCLDD, Deep_AML, HDML_CVPR19} \textit{supervised} in nature. 
This is undesirable, as obtaining manual annotations may either be infeasible, or may require domain-specific knowledge in some tasks. Particularly, for newer applications involving large datasets (for \eg, medical imaging requiring invasive procedures, 3D point cloud object datasets, pixel-level annotations for semantic segmentation), instances may be difficult to annotate.
While obtaining labels for large scale datasets is often infeasible, unlabeled data on the other hand, is ubiquitous and can provide richer information if exploited well. This motivates us to learn features or \textit{embeddings} in a completely \textit{unsupervised} manner, \ie, without using class labels.

\begin{figure}[t]
\centering
	\includegraphics[width=0.7\linewidth]{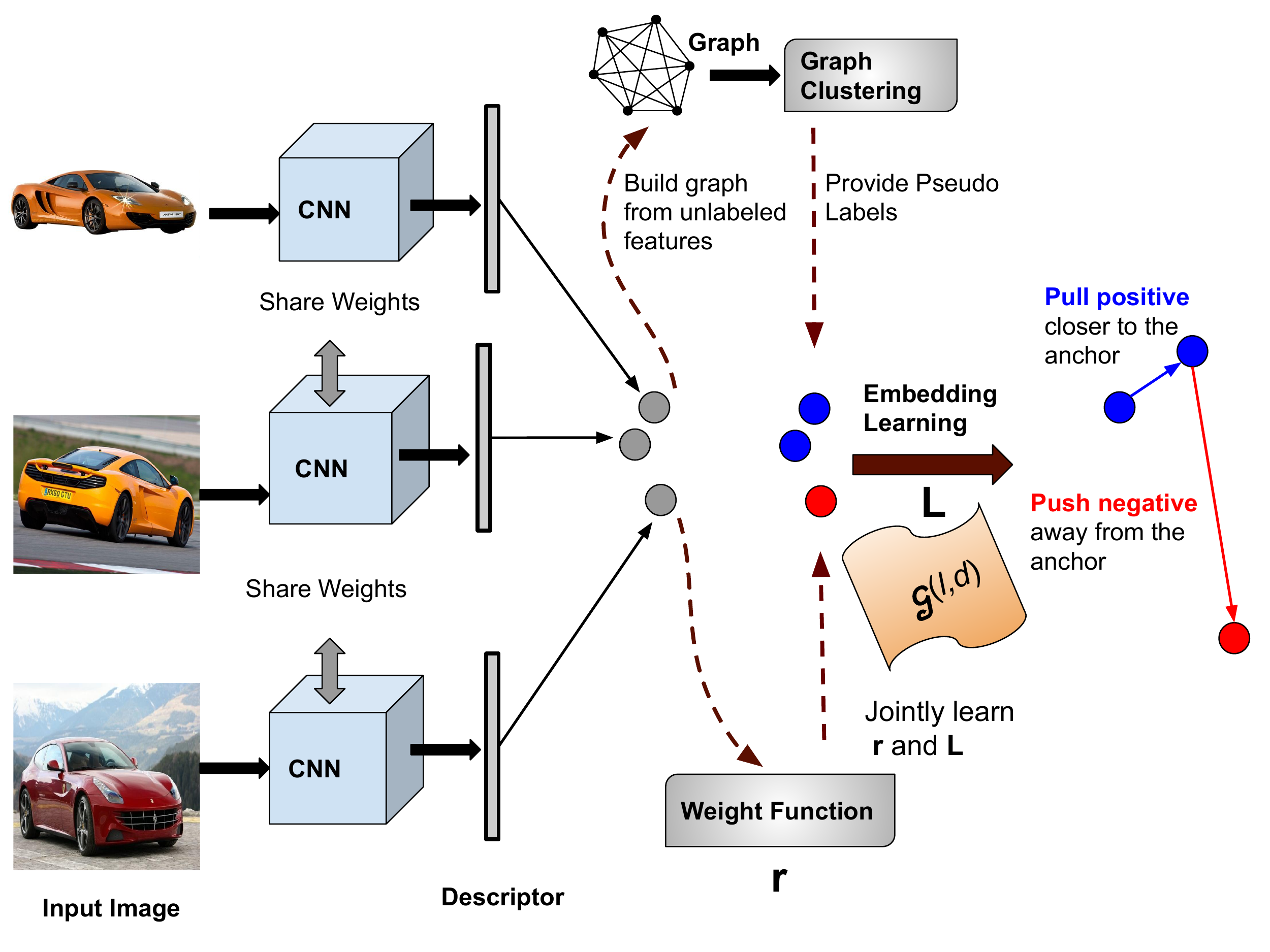}
    \caption{An illustration of the proposed embedding learning approach RPML. The input images belong to Cars196 dataset \cite{Cars196}. The figure is best viewed in color.}
    \label{framework_RPML}
\end{figure}

In this paper, we propose an unsupervised approach to learn an embedding that can achieve competitive results. Figure \ref{framework_RPML} illustrates our proposed approach. Particularly, we follow a \textit{triplets-based} metric learning paradigm \cite{LMNN, FaceNet, DRIFT} for providing \textit{weak supervision}. We believe, our study can establish new ways of performing \textit{self-training} and learning from unlabeled data. Furthermore, it leads up to a new, generic way to include information from unlabeled data, that could be used for general \textit{semi-supervised learning} in future.

As we do not have class labels, we use a graph-based clustering to obtain pseudo-labels, and form the triplet constraints. However, rather than naively using the triplets obtained from pseudo-labels, we propose to use a weight function that appropriately scales the losses associated with the triplets. Having the weighted losses for the triplets, we finally use a probabilistic notion to learn an embedding. In particular, we minimize the chances of each triplet violating a geometric angular constraint.

Additionally, to avoid the model from collapsing to a singularity and overfit to the training data, we impose constraints in the form of orthogonality on the parametric matrix of the learned embedding. This requires us to exploit Riemannian manifold based optimization techniques to learn the parameters of our approach. We jointly learn the parameters of the weight function and the embedding using optimization on a Riemannian product manifold. 

\section{Background}
This section provides a basic introduction to the following topics (relevant to our paper): i) Triplets-based Distance Metric Learning (DML) along with two relevant losses, ii) the graph-based Authority Ascent Shift (AAS) clustering method \cite{AAS}, used to obtain pseudo-labels for our proposed approach, and iii) basics of Riemannian optimization, which is used to learn the parameters of our method.

\subsection{Triplets-based Distance Metric Learning (DML)}
Let $\mathcal{X}=\{\Vec{x}_i\}_{i=1}^{|\mathcal{X}|}$ be a dataset of cardinality $|\mathcal{X}|$, with $\Vec{x}_i \in \mathbb{R}^d$ being the descriptor of sample $i$. For $\Vec{x}_i \in \mathbb{R}^d$, let $\Mat{L}^\top\Vec{x}_i \in \mathbb{R}^l$, denote its embedding learned by a projection matrix $\Mat{L}\in \mathbb{R}^{d \times l}$, such that $l \leq d$. $\Mat{L}$ facilitates dimensionality reduction, if $l<d$. Consider the function: $\delta^2_{\Mat{L}}(\Vec{x}_i,\Vec{x}_j)=(\Vec{x}_i-\Vec{x}_j)^\top\Mat{L}\Mat{L}^\top(\Vec{x}_i-\Vec{x}_j)$, for a pair of examples $\Vec{x}_i,\Vec{x}_j \in \mathbb{R}^d$. Here, $\delta^2_{\Mat{L}}(\Vec{x}_i,\Vec{x}_j)$ is essentially a squared-Mahalanobis distance. The goal of our work is to learn the parametric matrix $\Mat{L}$, given a set of \textit{triplet constraints} \cite{LMNN, SCML, DRIFT, FaceNet}: 
$\mathcal{T}_{\textrm{labeled}}= \{(\Vec{x}_i,\Vec{x}_i^+,\Vec{x}_i^-)\}_{i=1}^{|\mathcal{T}_{\textrm{labeled}}|}$. Here, $\Vec{x}_i^+$ is similar (or from same class) to $\Vec{x}_i$, and $\Vec{x}_i^-$ is dissimilar (or from different class) to both $\Vec{x}_i$ and $\Vec{x}_i^+$. $\Vec{x}_i$, $\Vec{x}_i^+$ and $\Vec{x}_i^-$ are respectively called as the \textit{anchor}, the \textit{positive} (or target) and the \textit{negative} (or impostor) respectively. The objective function to minimize the \textit{triplet-loss} can be expressed as follows:
\begin{equation}
\label{triplet_loss}
\begin{aligned}
J_{\textrm{triplet}}=\sum_{i=1}^{|\mathcal{T}_{\textrm{labeled}}|}[\delta^2_{\Mat{L}}(\Vec{x}_i,\Vec{x}_i^+)
-\delta^2_{\Mat{L}}(\Vec{x}_i,\Vec{x}_i^-)+\tau]_+.
\end{aligned}
\end{equation}
Here, (\ref{triplet_loss}) ensures that the distance between the pair $(\Vec{x}_i,\Vec{x}_i^-)$ is greater than distance between the pair $(\Vec{x}_i,\Vec{x}_i^+)$ by a margin $\tau>0$. $[z]_+=\textrm{max}(0,z), z\in \mathbb{R}$ denotes the hinge-loss function.
On the other hand, instead of constraining with respect to a distance-based margin $\tau$ as in (\ref{triplet_loss}), we can also constrain a triplet with respect to an angle. The objective function to minimize the \textit{angular loss} can be expressed as follows \cite{Angular_loss}:
\begin{equation}
\label{angular_loss}
\begin{aligned}
J_{\textrm{angular}}=\sum_{i=1}^{|\mathcal{T}_{\textrm{labeled}}|}[\delta^2_{\Mat{L}}(\Vec{x}_i,\Vec{x}_i^+)-4\textrm{ tan}^2\alpha \textrm{ } \delta^2_{\Mat{L}}(\Vec{x}_i^-,\Vec{x}_{i-avg})]_+.
\end{aligned}
\end{equation}
Here, $\Vec{x}_{i-avg}=(\Vec{x}_i+\Vec{x}_i^+)/2$, and $\alpha >0$ is a hyperparameter that corresponds to an upper bound on the angle at $\Vec{x}_i^-$ of the triplet $(\Vec{x}_i,\Vec{x}_i^+,\Vec{x}_i^-)$. In both (\ref{triplet_loss}) and (\ref{angular_loss}), the constraint set $\mathcal{T}_{\textrm{labeled}}$ is obtained using class labels. We however, do not make use of class labels to learn $\Mat{L}$.

\subsection{Obtaining pseudo-labels using Authority Ascent Shift (AAS) clustering}
We assume that the given dataset $\mathcal{X}$ is unlabeled. To compensate for the lack of class labels, we suggest obtaining pseudo-labels to form a triplet set. In our work, we choose the graph-based Authority Ascent Shift (AAS) clustering \cite{AAS} to obtain the pseudo-labels. Let, the AAS clustering be denoted by a function $c:\mathbb{R}^d\rightarrow\mathbb{Z}^+$ such that $c(\Vec{x}_i)$, a positive integer, denotes the \textit{pseudo-label} assigned to $\Vec{x}_i\in \mathbb{R}^d$. Briefly, AAS requires constructing a weighted graph with nodes representing the examples, and edges between nearest neighbors. Edge weights denote \textit{affinities} between examples. With $\omega$ denoting the stationary probability distribution of a random walker on the graph, the \textit{node relevancy} from node $i$ to node $j$ can be defined as \cite{AAS}:
\begin{equation}
   \label{node_relevancy}
    \psi(i,j)=d_i\Mat{P}_{ij}\exp(-\gamma(\nabla_{\omega}(i,j))^2).
\end{equation}
Here, $d_i$ is the out-degree of node $i$, $\Mat{P}_{ij}$ is the transition probability from node $i$ to node $j$, $\exp(.)$ is the exponential function, $\nabla_{\omega}(i,j)=[\omega(j)-\omega(i)]$ and $\gamma>0$ is a hyperparameter. The set of \textit{relevant neighbors} of node $i$ can be defined as:
\begin{equation}
   \label{relevant_neighbors}
    \mathcal{N}_\epsilon(i)=\left \{ j\in\mathcal{V}:\psi(i,j)> \epsilon \right \}\cup \left \{ i \right \}.
\end{equation}
Here, $\epsilon>0$ is a hyperparameter and $\mathcal{V}$ is the vertex set of the graph. \textit{Authority ascent} of a node $i$ can be performed by moving towards the node $j^*$ such that $j^*=\operatorname*{arg\,max}_{j \in \mathcal{N}_\epsilon(i)} \Mat{P}_{ij}\nabla_{\omega}(i,j)$. By subsequently performing authority ascent on neighboring nodes, we can associate a \textit{authority mode} \cite{AAS} to node $i$. Nodes sharing a common authority mode build a tree. Disjoint trees represent the distinct, arbitrary-shaped clusters present in the data \cite{AAS}. Using the clustering, we can obtain the set of \textit{triplets} as required.

\textbf{Motivation to use AAS:} AAS does not require the user to input the number of clusters present. It is very crucial in the unsupervised setting. More importantly, it is able to detect arbitrary-shaped clusters in the data, while being robust to noise and outliers. As AAS relies on a notion of geometric similarity, we further conjecture that AAS groups together far away \textit{semantically similar} examples. We believe that this is helpful in capturing \textit{intra-class variances} present in the data. For example, such variances occur in visual data due to minor pose, illumination or viewpoint differences. Capturing \textit{intra-class variances} is important, as also pointed out by Li \etal \cite{Cyclic_ECCV16}.

\subsection{Basics of Riemannian optimization}
\label{sec_riem_basics}
To learn the parameters of our proposed method, we make use of Riemannian manifold based optimization, which we believe, is an extensive study on its own. For a more detailed, formal study of the topic, we refer the interested reader to the book by Absil \etal \cite{AMS09}. However, to make the current text self-contained, we briefly explain the intuitions of a few underlying concepts relevant to our paper.

\begin{figure}[t]
\centering
	\includegraphics[width=0.3\linewidth]{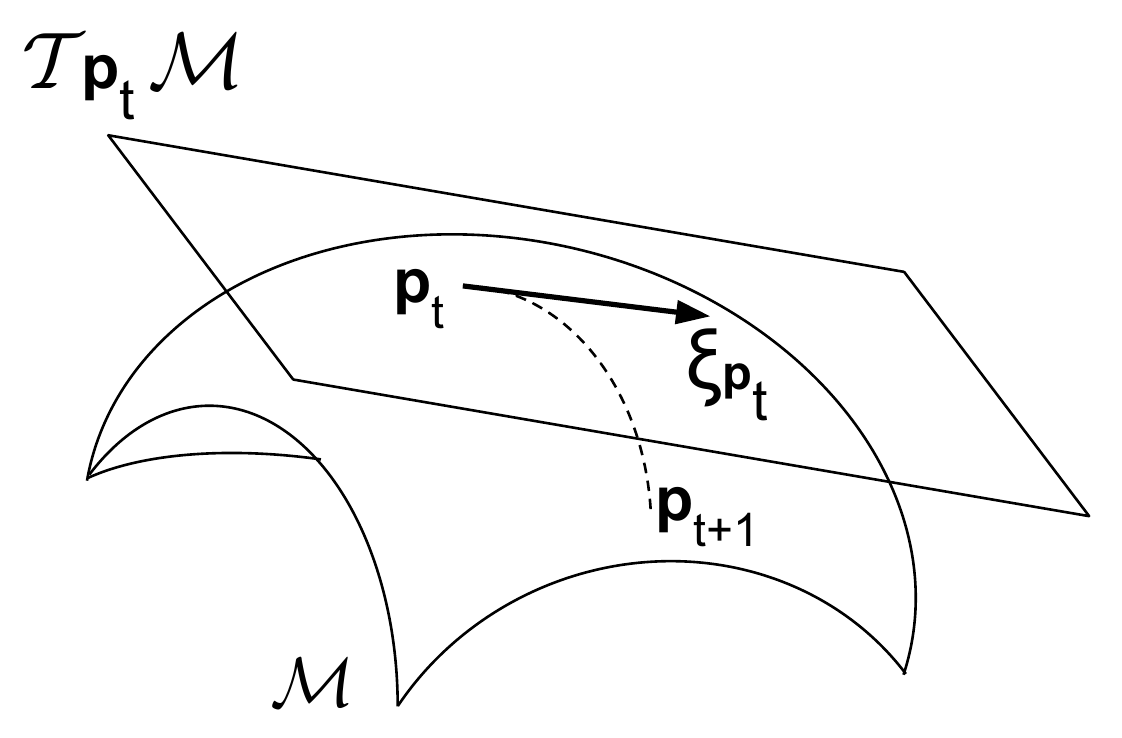}
    \caption{Riemannian manifold based optimization.}
    \label{manifold_opt}
\end{figure}

As shown in Figure \ref{manifold_opt}, a Riemannian manifold $\mathcal{M}$ is a non-Euclidean space that locally resembles an Euclidean space, and is equipped with an inner product on the \textit{tangent space} at each point $\Vec{p}_t\in\mathcal{M}$. The \textit{tangent space} $\mathcal{T}_{\Vec{p}_t}\mathcal{M}$ can be considered as a linearization of $\mathcal{M}$ at $\Vec{p}_t\in\mathcal{M}$.

Optimization methods like Riemannian Conjugate Gradient Descent (RCGD) can be performed by following a line-search algorithm on $\mathcal{M}$. Given a descent direction provided by the tangent vector $\xi_{\Vec{p}_t}\in \mathcal{T}_{\Vec{p}_t}\mathcal{M}$, line search can be performed by moving along a \textit{geodesic}, a smooth curve on the manifold $\mathcal{M}$ (dotted curve in Figure \ref{manifold_opt}). The update formula for the line-search algorithm can be given as:
\begin{equation}
\label{update_line_search}
\Vec{p}_{t+1}=\mathcal{R}_{\Vec{p}_{t}}(\eta_t\xi_{\Vec{p}_t}).
\end{equation}
Here, $\mathcal{R}_{\Vec{p}_{t}}(\eta_t\xi_{\Vec{p}_t})$ is called the \textit{retraction} operator, and $\eta_t>0$ is the step size. 

Let $f:\mathcal{M}\rightarrow \mathbb{R}$ be a smooth function. To minimize $f$, a gradient-descent algorithm can be obtained when direction of $\xi_{\Vec{p}_t}$ coincides with $-\textrm{grad }f(\Vec{p}_t)$, where $\textrm{grad }f(\Vec{p}_t)$ is the \textit{Riemannian gradient} at $\Vec{p}_t$. The \textit{Riemannian gradient} is defined as the unique element $\textrm{grad }f(\Vec{p}_t)\in \mathcal{T}_{\Vec{p}_t}\mathcal{M}$ that satisfies:
\begin{equation}
\label{dir_derivative}
D f(\Vec{p}_t)[\xi_{\Vec{p}_t}]=g(\textrm{grad }f(\Vec{p}_t),\xi_{\Vec{p}_t}).
\end{equation}
Here, $D f(\Vec{p}_t)[\xi_{\Vec{p}_t}]$ is the \textit{directional derivative} of $f(\Vec{p}_t)$ in the direction of $\xi_{\Vec{p}_t}$, and $g(\zeta_{\Vec{p}_t},\xi_{\Vec{p}_t})$ is the inner product between the tangent vectors $\zeta_{\Vec{p}_t}, \xi_{\Vec{p}_t}\in \mathcal{T}_{\Vec{p}_t}\mathcal{M}$.

Now, given a manifold $\mathcal{M}$ as shown in Figure \ref{manifold_opt}, a Lie group $G$ acts on $\mathcal{M}$, if it defines a mapping $h:\mathcal{M} \times G \rightarrow \mathcal{M}$. If the action of $G$ defines an equivalence relation $\sim$, then by the \textit{quotient manifold theorem} (Theorem 9.16) in Lee \etal \cite{lee2003smooth}, $\mathcal{M}_q = \mathcal{M}/\sim$ forms a smooth manifold, and is called a Riemannian \textit{quotient manifold}. Essentially, the quotient manifold $\mathcal{M}_q$ is the set of all equivalence classes, such that for a point $\Vec{x}\in \mathcal{M}$, the equivalence class is defined as: $[\Vec{x}]=\{\Vec{y}\in \mathcal{M}:\Vec{y} \sim \Vec{x}\}$. Thus, $\Vec{y}$ and $\Vec{x}$ represent the same point $[\Vec{x}]\in \mathcal{M}_q$.

As we will see later, our objective function (\ref{RPML}) displays an invariance property. In short, it means that for an objective function $f:\mathcal{M} \rightarrow \mathbb{R}$, and two points $\Vec{x}, \Vec{y} \in \mathcal{M}$ such that $\Vec{y} \sim \Vec{x}$, we get the same value of the objective, \ie, $f(\Vec{y})=f(\Vec{x})$. This could be detrimental to the optimization method. Hence, considering the quotient manifold theorem is important.

However, the quotient manifold $\mathcal{M}_q$ is an abstract manifold. Therefore, to provide a matrix representation of the abstract tangent space $\mathcal{T}_{[\Vec{x}]}\mathcal{M}_q$, we make use of $\mathcal{H}_{\Vec{x}}\mathcal{M}$. Here, $\mathcal{H}_{\Vec{x}}\mathcal{M}$ and $\mathcal{V}_{\Vec{x}}\mathcal{M}$ are called as the \textit{horizontal space} and \textit{vertical space} respectively, and are two complementary parts of the tangent space $\mathcal{T}_{\Vec{x}}\mathcal{M}$. 
The tangent vector $\xi_{\Vec{x}}\in \mathcal{H}_{\Vec{x}}\mathcal{M}$ is called the \textit{horizontal lift} of $\xi_{[\Vec{x}]}\in \mathcal{T}_{[\Vec{x}]}\mathcal{M}_q$.

The search spaces of the parameters of our approach (\ie, embedding and weight) are individual Riemannian manifolds. To jointly learn them we consider the product space of these manifolds, which is again a Riemannian manifold \cite{AMS09}. The operators discussed above, \ie, gradient and retraction, for this \textit{product manifold} can be defined as the Cartesian product of the individual components. However, the inner product is defined as the sum of the inner products of the individual components. 
Lastly, we discuss the following two manifolds that are relevant to our paper:
\begin{definition}{\textbf{The Stiefel manifold:}}
The (orthogonal) Stiefel manifold $\ST{d}{l}$ ($l \leq d$) \cite{AMS09}, is formed by the set of all orthonormal matrices of order $d \times l$, as follows:
\begin{equation*}
\ST{d}{l} \triangleq \{\Mat{L}\in \mathbb{R}^{d\times l}:\Mat{L}^\top\Mat{L}=\mathbf{I}_l \}
\end{equation*}
$\ST{d}{l}$ has a dimensionality of $dl-l(l+1)/2$. Here, $\mathbf{I}_l$ is the $l \times l$ identity matrix. The Riemannian metric is given as: $g(\xi_{\Mat{L}},\zeta_{\Mat{L}})=\Tr(\xi_{\Mat{L}}^\top\zeta_{\Mat{L}})$, for $\xi_{\Mat{L}},\zeta_{\Mat{L}}\in \mathcal{T}_{\Mat{L}}\ST{d}{l}$. $\Tr(.)$ is the trace operator.
\end{definition}

\begin{definition}{\textbf{The  Grassmann manifold:}}
The Grassmann manifold $\GRASS{d}{l}$ ($l \leq d$) \cite{AMS09}, is the collection of $l$-dimensional subspaces spanned by the columns of orthonormal matrices of order $d \times l$, and is defined as follows:
\begin{equation*}
\GRASS{d}{l} \triangleq \{\textrm{span}(\Mat{W}):\Mat{W}\in \mathbb{R}^{d\times l},\Mat{W}^\top\Mat{W}=\mathbf{I}_l \}
\end{equation*}
$\GRASS{d}{l}$ has a dimensionality of $(dl-l(l+1)/2)-l(l-1)/2$. $\GRASS{d}{l}$ is essentially a quotient manifold of $\ST{d}{l}$.
\end{definition}

\section{Proposed Method}

We now discuss our proposed embedding learning approach. Using the clustering function $c(.)$, we can generate a set of triplets: $\mathcal{T} =\{ (\Vec{x}_i,\Vec{x}_i^+,\Vec{x}_i^-) \}_{i=1}^{|\mathcal{T}|}$, each element of which consists of the following: i) $\Vec{x}_i$, an arbitrary example with a value $c(\Vec{x}_i)$
. ii) $\Vec{x}_i^+$, another arbitrary example with $c(\Vec{x}_i^+)=c(\Vec{x}_i)$. iii) $\Vec{x}_i^-$, such that $c(\Vec{x}_i^-) \neq c(\Vec{x}_i)$. The examples $\Vec{x}_i$, $\Vec{x}_i^+$ and $\Vec{x}_i^-$ are referred to as the \textit{anchor}, \textit{positive} and \textit{negative} respectively. Here, $|\mathcal{T}|$ is the number of triplets generated. Using $\mathcal{T}$, our goal is to learn $\Mat{L} \in \mathbb{R}^{d \times l} (l \leq d)$ in $\delta^2_{\Mat{L}}(\Vec{x}_i,\Vec{x}_j)=(\Vec{x}_i-\Vec{x}_j)^\top\Mat{L}\Mat{L}^\top(\Vec{x}_i-\Vec{x}_j)$.

Having obtained the \textit{constraint} set $\mathcal{T}$ of \textit{triplets}, we follow a \textit{metric learning} paradigm to learn $\Mat{L}$. 
Wang \etal \cite{Angular_loss} pointed that in the standard triplet loss as in (\ref{triplet_loss}), the gradients \wrt an example in a triplet consider only two of the examples at a time, and fail to capture the third. They propose an \textit{angular loss}  constraint as in (\ref{angular_loss}), that handles this issue by taking into account the angular geometry \wrt the negative example in the triplet. Thus, the gradients are computed considering all three examples of the triplet simultaneously. However, the loss in (\ref{angular_loss}) is non-smooth due to the hinge-loss function. Hence, to reap the benefits of manifold-based optimization (as we will see later), we suggest using a \textit{smooth} version of the \textit{angular constraint} as the \textit{metric loss} associated with a triplet $(\Vec{x}_i,\Vec{x}_i^+,\Vec{x}_i^-)$:
\begin{equation}
\label{metric_loss}
m(\Vec{x}_i,\Vec{x}_i^+,\Vec{x}_i^-)=\log(1+\exp(z(\Vec{x}_i,\Vec{x}_i^+,\Vec{x}_i^-))),
\end{equation}
where,
\begin{equation}
\label{z_i}
z(\Vec{x}_i,\Vec{x}_i^+,\Vec{x}_i^-)=\delta^2_{\Mat{L}}(\Vec{x}_i,\Vec{x}_i^+)-4\textrm{ tan}^2\alpha \textrm{ } \delta^2_{\Mat{L}}(\Vec{x}_i^-,\Vec{x}_{i-avg}).
\end{equation}
Here, $\Vec{x}_{i-avg}=\frac{\Vec{x}_i+\Vec{x}_i^+}{2}$. $\alpha >0$ corresponds to an upper bound on the angle at $\Vec{x}_i^-$ of the triplet $(\Vec{x}_i,\Vec{x}_i^+,\Vec{x}_i^-)$ \cite{Angular_loss}. Apart from the benefit of computing gradients \wrt three examples simultaneously, (\ref{metric_loss}) offers rotation and scale invariance. Furthermore, without a proper reference, it is not intuitive to tune a distance-based hyperparameter $\tau$ as in (\ref{triplet_loss}), whereas, it is much more intuitive to tune an angle $\alpha$ as present in (\ref{z_i}). 

Additionally, to scale the contribution of a triplet for learning our embedding, we also provide a weightage to the metric loss term defined in (\ref{metric_loss}). Specifically, we propose a \textit{\textbf{weighted loss}} as follows:
\begin{equation}
\label{wtd_loss}
f(\Vec{x}_i,\Vec{x}_i^+,\Vec{x}_i^-)=w(\Vec{x}_i,\Vec{x}_i^+,\Vec{x}_i^-)m(\Vec{x}_i,\Vec{x}_i^+,\Vec{x}_i^-).
\end{equation}
Here, a function $w:\mathbb{R}^d \times \mathbb{R}^d \times \mathbb{R}^d \rightarrow [0,1]$ is used to provide the \textit{weightage} given to the \textit{metric loss} associated with a triplet $(\Vec{x}_i,\Vec{x}_i^+,\Vec{x}_i^-)$, and we define it as follows:
\begin{equation}
\label{wt}
    w(\Vec{x}_i,\Vec{x}_i^+,\Vec{x}_i^-)=\frac{1}{1+\exp(-\Vec{r}^\top\Vec{x}_{i-cat})},
\end{equation}
where, $\Vec{x}_{i-cat}=[\Vec{x}_{i-avg}^\top ,(\Vec{x}_{i}^-)^\top]^\top$ and $\Vec{r}\in\mathbb{R}^{2d}$ represents the parameter of the weight function. Intuitively, we collectively represent the triplet as a single example, and let that representation decide the weightage that should be given to the \textit{metric loss} term for the triplet.
Concatenating examples within a triplet as a single vector can help to capture specific relationships among them. For example, Duan \etal \cite{Deep_AML} have concatenated the examples in a triplet to generate a synthetic \textit{negative} with respect to the \textit{anchor}. In our case, we map the concatenation of the examples in a triplet to a \textit{confidence} value.

We now propose a novel probabilistic objective to learn an embedding. Given the set of parameters $\Mat{L}$ and $\Vec{r}$ of our model, we associate a probability to a triplet $(\Vec{x}_i,\Vec{x}_i^+,\Vec{x}_i^-)$, defined as follows:
\begin{equation}
\label{prob_triplet}
p_i=\frac{1}{1+\exp(f_i)}.
\end{equation}
Here, $f_i$ denotes $f(\Vec{x}_i,\Vec{x}_i^+,\Vec{x}_i^-)$ in (\ref{wtd_loss}). We know that for the triplet $(\Vec{x}_i,\Vec{x}_i^+,\Vec{x}_i^-)$ to satisfy the \textit{angular loss constraint}, we can \textit{minimize} $f_i$. In (\ref{prob_triplet}), minimizing $f_i$ ensures maximizing $p_i$. Hence, $p_i$ intuitively represents the likelihood of the triplet \textit{satisfying the angular constraint}. Now, let $J_i=-\log p_i=\log(1+\exp(f_i))$, denote the negative log likelihood associated with the triplet $(\Vec{x}_i,\Vec{x}_i^+,\Vec{x}_i^-)$. We propose to use Maximum Likelihood Estimation (MLE) to learn our parameters, \ie, we minimize the sum of the negative log likelihood over all the triplets, as follows:
\begin{equation}
\label{RPML}
\min_{\Vec{r},\Mat{L}} \mathcal{L}(\Vec{r},\Mat{L})=\sum_{i=1}^{|\mathcal{T}|}J_i=\sum_{(\Vec{x}_i,\Vec{x}_i^+,\Vec{x}_i^-)\in \mathcal{T}} \log(1+\exp(f_i)).
\end{equation}

However, without any regularizer, our model may collapse to a singularity. To avoid this, we enforce the \textit{orthogonality constraint} $\Mat{L}^\top\Mat{L}=\mathbf{I}_l\in\mathbb{R}^{l\times l}$. Adding this constraint has the following additional benefits \cite{MDMLCLDD}: i) minimizing the difference in performance for frequent and less frequent classes, ii) avoiding overfitting, and iii) obtaining a better embedding using a small number of projection vectors.

The matrix $\Mat{L}$ naturally lies on a orthogonal Stiefel manifold $\ST{d}{l}$ \cite{AMS09}. However, the objective of (\ref{RPML}) is invariant to the right action of the orthogonal group $\ORTHO{l}=\{\Mat{B}\in \mathbb{R}^{l\times l}:\Mat{B}^\top\Mat{B}=\mathbf{I}_l\}$, \ie, for $\Mat{B}\in \ORTHO{l}$, we have $\mathcal{L}(\Vec{r},\Mat{L})=\mathcal{L}(\Vec{r},\Mat{L}\Mat{B})$. To jointly learn the parameters $\Vec{r}$ and $\Mat{L}$, we can constrain the optimization problem in (\ref{RPML}) on the following product manifold:
\begin{equation}
\label{prod_manifold}
    \mathcal{M}_{p}\triangleq\mathbb{R}^{2d}\times \GRASS{d}{l}.
\end{equation}
Here, $\GRASS{d}{l}$ is the Grassmann manifold \cite{AMS09}, which is the quotient of $\ST{d}{l}$ with the equivalence class being $[\Mat{L}]\triangleq\{\Mat{L}\Mat{B}, \Mat{L}\in \ST{d}{l}, \Mat{B}\in \ORTHO{l}\}$. $\mathcal{M}_{p}$ can be given the structure of a \textit{Riemannian manifold} using product topology \cite{AMS09}, and it has a dimensionality of $2d+(dl-l(l+1)/2)-l(l-1)/2$. We call our proposed approach defined in (\ref{RPML}) as \textbf{\textit{Reweighted Probabilistic unsupervised eMbedding Learning (RPML)}}. Figure \ref{framework_RPML} illustrates the RPML approach.

We can learn the parameters of RPML using Riemannian Conjugate Gradient Descent (RCGD), for which we require the Euclidean gradients $\nabla_{\Vec{r}}\mathcal{L}({\Vec{r}},{\Mat{L}})=\sum_{i=1}^{|\mathcal{T}|}\nabla_{\Vec{r}}J_i$ and $\nabla_{\Mat{L}}\mathcal{L}({\Vec{r}},{\Mat{L}})=\sum_{i=1}^{|\mathcal{T}|}\nabla_{\Mat{L}}J_i$. Here,
\begin{equation}
\nabla_{\Vec{r}}J_i=g_ih_im_i\Vec{x}_{i-cat},
\end{equation}
\begin{equation}
\nabla_{\Mat{L}}J_i=2g_iw_i \beta_i[\Vec{\delta}_{ap}\Vec{\delta}_{ap}^\top-4\textrm{ tan}^2\alpha \Vec{\delta}_{nm}\Vec{\delta}_{nm}^\top]\Mat{L},
\end{equation}
$g_i=\exp(f_i)/(1+\exp(f_i))$, $h_i=\exp(-\Vec{r}^\top\Vec{x}_{i-cat})/(1+\exp(-\Vec{r}^\top\Vec{x}_{i-cat}))^2$, $m_i$ denotes $m(\Vec{x}_i,\Vec{x}_i^+,\Vec{x}_i^-)$ in (\ref{metric_loss}). Also, $w_i$ denotes $w(\Vec{x}_i,\Vec{x}_i^+,\Vec{x}_i^-)$ in (\ref{wt}), $\beta_i=\exp(z_i)/(1+\exp(z_i))$, $z_i$ denotes $z(\Vec{x}_i,\Vec{x}_i^+,\Vec{x}_i^-)$ in (\ref{z_i}), $\Vec{\delta}_{ap}=\Vec{x}_i-\Vec{x}_i^+$ and $\Vec{\delta}_{nm}=\Vec{x}_i^--\Vec{x}_{i-avg}$. Given the Euclidean gradients, we can use a standard toolbox like Manopt \cite{manopt} to perform RCGD.

\subsection{An efficient algorithm for the proposed RPML method}

We now propose an efficient matrix-vector based algorithm for RPML. Given $\mathcal{T} =\{ (\Vec{x}_i,\Vec{x}_i^+,\Vec{x}_i^-) \}_{i=1}^{|\mathcal{T}|}$, we can construct matrices $\Mat{A}=[\Vec{a}_1 \cdots \Vec{a}_i \cdots \Vec{a}_{|\mathcal{T}|}] \in \mathbb{R}^{d \times |\mathcal{T}|}$ and $\Mat{C}=[\Vec{c}_1 \cdots \Vec{c}_i \cdots \Vec{c}_{|\mathcal{T}|}] \in \mathbb{R}^{2d \times |\mathcal{T}|}$, where $\Vec{a}_i=(\Vec{x}_i+\Vec{x}_i^+)/2$ and $\Vec{c}_i=[ \Vec{a}_i^\top, (\Vec{x}_i^-)^\top ]^\top$. We can denote a vector $\boldsymbol \rho \in \mathbb{R}^{|\mathcal{T}|}$, each component of which is computed as: $\rho_i=1+\exp(-\Vec{r}^\top\Vec{c}_i)$. Denoting $\Vec{e}_{|\mathcal{T}|} \in \mathbb{R}^{|\mathcal{T}|}$ as a vector of all ones, we can compute the following vector: $\Vec{w}=\Vec{e}_{|\mathcal{T}|} \oslash  \boldsymbol \rho$, where $\Vec{w} \in \mathbb{R}^{|\mathcal{T}|}$ and $\oslash$ denotes the component-wise division operator.

We can also construct the following matrices $\Mat{P}=[\Vec{p}_1 \cdots \Vec{p}_i \cdots \Vec{p}_{|\mathcal{T}|}] \in \mathbb{R}^{d \times |\mathcal{T}|}$ and $\Mat{Q}=[\Vec{q}_1 \cdots \Vec{q}_i \cdots \Vec{q}_{|\mathcal{T}|}] \in \mathbb{R}^{d \times |\mathcal{T}|}$, where $\Vec{p}_i=\Vec{x}_i-\Vec{x}_i^+$ and $\Vec{q}_i= \Vec{x}_i^--\Vec{a}_i$. Denoting $\Vec{e}_l \in \mathbb{R}^l$ as a vector of all ones, we can compute vectors $\Vec{z},\Vec{m},\Vec{f}\in \mathbb{R}^{|\mathcal{T}|}$. Let, $\Vec{z}=(\Mat{L}^\top\Mat{P} \odot \Mat{L}^\top\Mat{P})^\top\Vec{e}_l-4\textrm{ tan}^2\alpha(\Mat{L}^\top\Mat{Q} \odot \Mat{L}^\top\Mat{Q})^\top\Vec{e}_l$, and $\Vec{m}$ be a vector with its $i$-th component computed as: $m_i=\log(1+\exp(z_i))$. Then, $\Vec{f}=\Vec{w}\odot\Vec{m}$, where $\odot$ is the Hadamard product operator. We can denote $ \Vec{\tilde{f}} \in \mathbb{R}^{|\mathcal{T}|}$, each component of which can be computed as:
\begin{equation}
\label{ftilde_vec}
\tilde{f}_i=\log(1+\exp(f_i)),
\end{equation}
where $f_i$ denotes the $i$-th component of $\Vec{f}$. Then, the objective of RPML can be expressed as:
\begin{equation}
\label{RPML_obj_vec}
\mathcal{L}(\Vec{r},\Mat{L})=\Tr(\textrm{diag}(\Vec{\tilde{f}})).
\end{equation}
Here, $\Tr(.)$ is the trace operator and $\textrm{diag}(\Vec{\tilde{f}})$ is a diagonal matrix.

Now, let $\Vec{g},\Vec{h}, \Vec{\beta}\in \mathbb{R}^{|\mathcal{T}|}$ be vectors, components of which are computed as: $g_i=\exp(f_i)/(1+\exp(f_i))$, $h_i=\exp(-\Vec{r}^\top\Vec{c}_i)/(1+\exp(-\Vec{r}^\top\Vec{c}_i))^2$ and $\beta_i=\exp(z_i)/(1+\exp(z_i))$. We can construct the following matrices: $\Mat{\tilde{C}}=[\Vec{\tilde{c}}_1 \cdots \Vec{\tilde{c}}_i \cdots \Vec{\tilde{c}}_{|\mathcal{T}|}] \in \mathbb{R}^{2d \times |\mathcal{T}|}$, $\Mat{\tilde{P}}=[\Vec{\tilde{p}}_1 \cdots \Vec{\tilde{p}}_i \cdots \Vec{\tilde{p}}_{|\mathcal{T}|}] \in \mathbb{R}^{d \times |\mathcal{T}|}$ and $\Mat{\tilde{Q}}=[\Vec{\tilde{q}}_1 \cdots \Vec{\tilde{q}}_i \cdots \Vec{\tilde{q}}_{|\mathcal{T}|}] \in \mathbb{R}^{d \times |\mathcal{T}|}$ as follows:
\begin{equation}
\label{C_tilde_matrix}
\tilde{c}_{ik}=[\Vec{g}^\top \odot \Vec{h}^\top \odot \Vec{m}^\top]_ic_{ik},
\end{equation}
\begin{equation}
\label{P_tilde_matrix}
\tilde{p}_{ik}=[2\Vec{g}^\top \odot \Vec{w}^\top \odot \Vec{\beta}^\top]_ip_{ik},
\end{equation}
\begin{equation}
\label{Q_tilde_matrix}
\tilde{q}_{ik}=[2\Vec{g}^\top \odot \Vec{w}^\top \odot \Vec{\beta}^\top]_iq_{ik}.
\end{equation}
Here, $c_{ik},\tilde{c}_{ik},p_{ik},\tilde{p}_{ik},q_{ik}$ and $\tilde{q}_{ik}$ are the $k$-th components of vectors $\Vec{c}_i,\Vec{\tilde{c}}_i,\Vec{p}_i,\Vec{\tilde{p}}_i,\Vec{q}_i$ and $\Vec{\tilde{q}}_i$ respectively. Also, $[\Vec{g}^\top \odot \Vec{h}^\top \odot \Vec{m}^\top]_i$ and $ [2\Vec{g}^\top \odot \Vec{w}^\top \odot \Vec{\beta}^\top]_i$ denote the $i$-th components of the respective vectors. Now, we can compute the Euclidean gradients as follows:
\begin{equation}
\label{grad_r}
\nabla_\Vec{r}\mathcal{L}(\Vec{r},\Mat{L})=\Mat{\tilde{C}}\Vec{e}_{|\mathcal{T}|},
\end{equation}
\begin{equation}
\label{grad_L}
\nabla_\Mat{L}\mathcal{L}(\Vec{r},\Mat{L})=(\Mat{\tilde{P}}\Mat{P}^\top-4\textrm{ tan}^2\alpha\Mat{\tilde{Q}}\Mat{Q}^\top)\Mat{L}.
\end{equation}

We can utilize the product topology of (\ref{prod_manifold}) and perform a Riemannian Conjugate Gradient Descent (RCGD) method \cite{AMS09} to learn the parameters of RPML. Let $\nabla_\Vec{r}\mathcal{L}$ and $\nabla_\Mat{L}\mathcal{L}$ be shorthand notations for (\ref{grad_r}) and (\ref{grad_L}). The Riemannian gradient is obtained as $\textrm{grad}_{\Mat{L}}= \nabla_{\Mat{L}}\mathcal{L}-\Mat{L}_t\Mat{L}_t^\top\nabla_{\Mat{L}}\mathcal{L}$. For $\Mat{L}_t\in \GRASS{d}{l}$ and the tangent vector $\xi_{\Mat{L}_t}=-\eta \textrm{ grad}_{\Mat{L}}$, the retraction $\mathcal{R}_{\Mat{L}_t}(\xi_{\Mat{L}_t})$ can be obtained as: $\Mat{U}\Mat{V}^\top$, where $[\Mat{U},\Mat{\Lambda},\Mat{V}]=\textrm{SVD}(\Mat{L}_t+\xi_{\Mat{L}_t})$. Hence, we can provide the following update steps for the parameters $\Vec{r}$ and $\Mat{L}$ (with update step for $\Vec{r}$ being the usual Euclidean gradient update):
\begin{equation}
\label{update_r}
\Vec{r}_{t+1}=\Vec{r}_t-\eta  \nabla_\Vec{r}\mathcal{L},
\end{equation}
\begin{equation}
\label{update_L}
\Mat{L}_{t+1}=\Mat{U}\Mat{V}^\top.
\end{equation}
Here, $\Mat{U}$ and $\Mat{V}$ are matrices obtained using a SVD such that $\Mat{U}\Mat{\Lambda}\Mat{V}^\top=\Mat{L}_t-\eta \nabla_{\Mat{L}}\mathcal{L}+\eta \Mat{L}_t\Mat{L}_t^\top\nabla_{\Mat{L}}\mathcal{L}$. $\Vec{r}_t$ and $\Mat{L}_t$ denote the parameters at the $t$-th iteration, and $\eta >0$ is the learning rate. $\Mat{\Lambda}$ is a diagonal matrix. 
The proposed algorithm for RPML has been summarized in Algorithm \ref{alg_RPML}.

\begin{algorithm}
\caption{Reweighted Probabilistic unsupervised eMbedding Learning (RPML)}
\label{alg_RPML}
\begin{algorithmic}[1]
\Require Unlabeled data $\mathcal{X}=\{\Vec{x}_j\in \mathbb{R}^d\}_{j=1}^{|\mathcal{X}|}$; $c:\mathbb{R}^d\rightarrow\mathbb{Z}^+$; $\alpha, maxiter>0$.
\State Initialize $\Vec{r}, \Mat{L}$;
\State Generate $\mathcal{T} =\{ (\Vec{x}_i,\Vec{x}_i^+,\Vec{x}_i^-) \}_{i=1}^{|\mathcal{T}|}$ using $c(.)$ on $\mathcal{X}$.
\State Construct $\Mat{A},\Mat{C},\Mat{P}$ and $\Mat{Q}$.
\State Compute $\Vec{w}, \Vec{z}, \Vec{m}, \Vec{f}, \Vec{g}, \Vec{h}$ and $\Vec{\beta}$.
\For{$iter=1 $ \textbf{to} $maxiter$}
\State $[\Vec{r}, \Mat{L}] \gets  \textrm{Perform RCGD}.$ \Comment{Using (\ref{grad_r}), (\ref{grad_L})}
\EndFor\label{for_iter}
\State \textbf{return} $\Mat{L}$
\end{algorithmic}
\end{algorithm}

\subsection{RPML using an alternative weight function (RPMLv1)}
We now discuss another alternative to the weight function that we defined in (\ref{wt}). We propose to express our alternative weight function as:
\begin{equation}
\label{wt1}
    w(\Vec{x}_i,\Vec{x}_i^+,\Vec{x}_i^-)=w_i=(w_i^++w_i^-)/2,
\end{equation}
where
\begin{equation}
\label{w_+}
    w_i^+=\frac{1}{1+\exp(-\Vec{x}_i^\top\Mat{R}\Mat{R}^\top\Vec{x}_i^+)},
\end{equation}
and
\begin{equation}
\label{w_-}
    w_i^-=1-\frac{1}{1+\exp(-\Vec{x}_{i-avg}^\top\Mat{R}\Mat{R}^\top\Vec{x}_i^-)}.
\end{equation}
Here, $\Mat{R}\in\mathbb{R}^{d\times l}$ is a matrix that is used to project the examples to a $l$-dimensional space (we fix it equal to the dimensionality of the embedding space, but it could be different). $\Vec{x}_i^\top\Mat{R}\Mat{R}^\top\Vec{x}_i^+$ represents the \textit{bilinear similarity} of the \textit{anchor} $\Vec{x}_i$, \wrt the \textit{positive} $\Vec{x}_i^+$, in the space projected by $\Mat{R}$. A lower value of similarity indicates that $\Vec{x}_i$ and $\Vec{x}_i^+$ should not have been grouped together by the clustering, and hence a lower weightage $w_i^+$ should be given to the loss associated with this triplet.

$\Vec{x}_{i-avg}^\top\Mat{R}\Mat{R}^\top\Vec{x}_i^-$ represents the \textit{bilinear similarity} of the \textit{negative} $\Vec{x}_i^-$, \wrt the average representation $\Vec{x}_{i-avg}$ of the similar pair (anchor, positive) in the triplet, in the space projected by $\Mat{R}$. A higher similarity indicates that $\Vec{x}_i^-$ is similar to the anchor (and hence positive), and all three of them should have been grouped together by the clustering. Hence, a lower weightage $w_i^-$ should be given to the loss term associated with this triplet while learning the embedding, which explains the subtraction from 1 in (\ref{w_-}).

We call the version of RPML obtained using this weight function as \textbf{RPMLv1}. Please note that RPMLv1 requires replacing the parameter $\Vec{r}$ in (\ref{RPML}) by $\Mat{R}$. To learn the parameters in RPMLv1, the product manifold is changed as follows: $\mathcal{M}_{p}\triangleq\mathbb{R}^{d \times l}\times \GRASS{d}{l}$. While the Euclidean gradient \wrt $\Mat{L}$ is similar as in RPML, we provide the form of $\nabla_{\Mat{R}}\mathcal{L}({\Mat{R}},{\Mat{L}})=\sum_{i=1}^{|\mathcal{T}|}\nabla_{\Mat{R}}J_i$.
Here,
\begin{equation}
\nabla_{\Mat{R}}J_i=\frac{1}{2}[g_ih_i^+m_i\Vec{\Delta}_{ap}-g_ih_i^-m_i\Vec{\Delta}_{mn}]\Mat{R},
\end{equation}
$J_i$ is the negative log-likelihood for a triplet, as defined earlier. $g_i=\exp(f_i)/(1+\exp(f_i))$, $h_i^+=\exp(-\Vec{x}_i^\top\Mat{R}\Mat{R}^\top\Vec{x}_i^+)/(1+\exp(-\Vec{x}_i^\top\Mat{R}\Mat{R}^\top\Vec{x}_i^+))^2$, $h_i^-=\exp(-\Vec{x}_{i-avg}^\top\Mat{R}\Mat{R}^\top\Vec{x}_i^-)/(1+\exp(-\Vec{x}_{i-avg}^\top\Mat{R}\Mat{R}^\top\Vec{x}_i^-))^2$, $m_i$ denotes $m(\Vec{x}_i,\Vec{x}_i^+,\Vec{x}_i^-)$ in (\ref{metric_loss}), $\Vec{\Delta}_{ap}=(\Vec{x}_i\Vec{x}_i^{+^\top}+\Vec{x}_i^+\Vec{x}_i^\top)$ and $\Vec{\Delta}_{mn}=(\Vec{x}_{i-avg}\Vec{x}_i^{-^\top}+\Vec{x}_i^-\Vec{x}_{i-avg}^\top)$.

\section{Related Work}
\label{rel_work}

\textbf{Supervised methods:} For challenging settings like \textit{zero-shot learning} \cite{ZSL_good_bad_ugly, schonfeld2019generalized}, \textit{extreme classification} \cite{yen2016pd,prabhu2014fastxml}, \textit{few-shot learning} \cite{sun2019meta, wertheimer2019few, triantafillou2017few}, and \textit{Fine-Grained Visual Categorization} (FGVC) \cite{qian2015fine}, \textit{metric learning} \cite{Bellet2015, survey2017} is preferred over standard classification based models. This is because metric learning has the ability to learn \textit{generic} notions of similarities, as opposed to \textit{class-specific} concepts. This is particularly crucial in problems like zero-shot learning where the test data belong to categories not seen during training. Metric learning implicitly learns an \textit{embedding} that groups similar examples, while moving away dissimilar ones. Alternately, an embedding commonly induces a metric in the learned space.

The literature on metric learning is rich, and we refer the interested reader to the surveys by Bellet \etal \cite{Bellet2015} and Lu \etal \cite{survey2017}. To form constraints for \textit{weak supervision}, commonly used metric learning approaches employ \textit{pairs} \cite{AML, MDMLCLDD, LR-GMML}, \textit{triplets} \cite{DRIFT, FaceNet} etc. 
The role of quality of the constraints in the training convergence has been pointed out lately \cite{FaceNet, N_pair}. Schroff \etal \cite{FaceNet} discusses the selection of \textit{semi-hard} examples for metric learning. Sohn \etal \cite{N_pair} proposes to move away multiple negative examples. Duan \etal \cite{Deep_AML} and Chen \etal \cite{AML} introduce \textit{adversarial} \cite{GAN} constraints for metric learning. Despite their merits, all these approaches depend on a huge number of labeled examples, which may be infeasible to obtain in many applications, as already discussed.

\textbf{Unsupervised methods:}
Despite the availability of a plethora of \textit{supervised} approaches that make use of class labels to generate constraints for metric or embedding learning, the number of \textit{unsupervised} approaches is quite limited. Classically, learning of embeddings in an unsupervised manner has been studied in the context of \textit{dimensionality reduction} or \textit{manifold learning} \cite{MDS, ISOMAP, LPP, NPE}. However, they either do not generalize for out-of-sample data, or are not specifically aimed at metric learning. \textit{Diffusion processes} \cite{donoser2013diffusion,bai2017ensemble,iscen2017efficient} capture the intrinsic manifold structure of the data and propagate affinities through a pairwise similarity matrix by random-walk steps.

Many existing unsupervised feature learning approaches that have been proposed recently \cite{ExemplarCNN_TPAMI16, Cyclic_ECCV16, DeepCluster_ECCV18, NCE_CVPR18} ignore the basic \textit{semantic relationships} among the examples. They are not particularly designed for metric learning. As such, they do not perform well in difficult settings like \textit{zero-shot learning}, where test examples belong to unseen \textit{semantic categories}. Recently, Iscen \etal \cite{MOM} proposed an approach for metric learning by mining \textit{hard} positives and negatives in an unsupervised manner. Ye \etal \cite{InvariantSpread_CVPR19} recently proposed another state-of-the-art method based on softmax embeddings.

\section{Empirical Evaluations}

\subsection{Quantitative results for fine-grained categorization}
\paragraph{Datasets and Evaluation protocol:} Following standard deep metric learning protocol in the zero-shot setting \cite{InvariantSpread_CVPR19}, we conduct experiments using a stochastic version of our RPML approach (mini-batch size of 120), for the task of fine-grained visual categorization on the Caltech-UCSD Birds 200 (CUB) \cite{CUB}, Cars-196 \cite{Cars196} and Stanford Online Products (SOP) \cite{Lifted_structure} datasets. CUB consists of 200 species of images of birds with first 100 species (5864 examples) for training and remaining (5924 examples) for testing. Cars-196 consists of images of cars from 196 models. We used first 98 models (8054 images) for training and remaining (8131) for testing. For the SOP dataset, that consists of 22634 classes with 120053 images of products, we used first 11318 classes (59551 images) for training and remaining 11316 classes (60502 images) for testing.

We used GoogLeNet \cite{GoogLeNet} pretrained on ImageNet \cite{ImageNet2015}, as the backbone CNN, using the MatConvNet \cite{MatConvNet} tool. The initial features are formed by the Regional Maximum Activation of Convolutions (R-MAC) \cite{RMAC_ICLR16} right before the average pool layer (following the MOM approach \cite{MOM}), aggregated over three input scales ($512$, $512/\sqrt{2}$, $256$).

\paragraph{Hyperparameters:} For all datasets, we set $\alpha=45^\circ$ (following Wang \etal \cite{Angular_loss}). For obtaining the initial clustering, we used a 50 nearest neighbor graph for the AAS clustering approach \cite{AAS}. AAS has two important parameters: $\gamma$ in (\ref{node_relevancy}) and $\epsilon$ in (\ref{relevant_neighbors}). As $\gamma$ is merely a scale parameter, we set it to $10^2$. Using the t-SNE embedding of training data, we observed that setting $\epsilon$ to a lower value easily merges nearby examples into the same cluster. This leads to huge clusters with large spread, which is undesirable. On the other hand, setting a higher value leads to difficulty in merging examples into clusters. Hence for all datasets, we arbitrarily set $\epsilon=0.65$, a value that is neither too high, nor too low. 
\begin{table*}[t]
\centering
\caption{Comparison against deep unsupervised approaches on three benchmark fine-grained datasets. The best performing method is shown in \textbf{bold black}, while the second best method is shown in \textbf{\textcolor{blue}{bold blue}}.}
\label{vs_deep_all}
\resizebox{0.9\linewidth}{!}{%
\begin{tabular}{|c|c|ccccc|ccccc|cccc|}
\hline
\multicolumn{2}{|c|}{\textbf{Dataset}}       & \multicolumn{5}{c|}{\textbf{CUB 200} \cite{CUB}}                                         & \multicolumn{5}{c|}{\textbf{Cars 196} \cite{Cars196}}                                        & \multicolumn{4}{c|}{\textbf{Stanford Online Products} \cite{Lifted_structure}}         \\ \hline
\textbf{Method}      & \textbf{Class Labels} & \textbf{NMI}  & \textbf{R@1}  & \textbf{R@2}  & \textbf{R@4}  & \textbf{R@8}  & \textbf{NMI}  & \textbf{R@1}  & \textbf{R@2}  & \textbf{R@4}  & \textbf{R@8}  & \textbf{NMI}  & \textbf{R@1}  & \textbf{R@10} & \textbf{R@100} \\ \hline
Intitial (Random)    & No                    & 34.6          & 31.5          & 42.4          & 55.0          & 67.0          & 23.4          & 22.7          & 31.7          & 42.7          & 54.8          & 79.8          & 25.4          & 35.6          & 48.5           \\ \hline
Exemplar \cite{ExemplarCNN_TPAMI16}             & No                    & 45.0          & 38.2          & 50.3          & 62.8          & 75.0          & 35.4          & 36.5          & 48.1          & 59.2          & 71.0          & 85.0          & 45.0          & 60.3          & 75.2           \\
NCE \cite{NCE_CVPR18}                 & No                    & 45.1          & 39.2          & 51.4          & 63.7          & 75.8          & 35.6          & 37.5          & 48.7          & 59.8          & 71.5          & \textbf{\textcolor{blue}{85.8}} & 46.6          & 62.3          & 76.8           \\
DeepCluster \cite{DeepCluster_ECCV18}         & No                    & 53.0          & 42.9          & 54.1          & 65.6          & 76.2          & 38.5          & 32.6          & 43.8          & 57.0          & 69.5          & 82.8          & 34.6          & 52.6          & 66.8           \\
MOM \cite{MOM}                 & No                    & \textbf{\textcolor{blue}{55.0}} & \textbf{\textcolor{blue}{45.3}} & \textbf{\textcolor{blue}{57.8}} & \textbf{\textcolor{blue}{68.6}} & 78.4          & \textbf{\textcolor{blue}{38.6}}          & 35.5          & 48.2          & 60.6          & 72.4          & 84.4          & 43.3          & 57.2          & 73.2           \\
Invariant \cite{InvariantSpread_CVPR19}                 & No                    & \textbf{55.4} & \textbf{46.2} & \textbf{59.0} & \textbf{70.1} & \textbf{80.2}          & 35.8          & \textbf{\textcolor{blue}{41.3}}          & \textbf{\textcolor{blue}{52.3}}          & \textbf{\textcolor{blue}{63.6}}          & \textbf{\textcolor{blue}{74.9}}          & \textbf{86.0}          & \textbf{48.9}          & \textbf{64.0}          & 78.0           \\\hline
\textbf{Ours} & No                    & 53.1          & 43.7          & 56.5          & 68.3          & \textbf{\textcolor{blue}{79.4}} & \textbf{40.0} & \textbf{44.6} & \textbf{56.9} & \textbf{68.6} & \textbf{79.5} & 82.5          & \textbf{\textcolor{blue}{47.5}} & \textbf{\textcolor{blue}{63.5}} & \textbf{78.1}  \\ \hline
\end{tabular}%
}
\end{table*}

\paragraph{Evaluation metrics:} After learning the parameters of our method using the training data, we project the test data in the embedding space, where we perform clustering and retrieval on the test data. To compare the methods, we perform $k$-means clustering by setting the value of $k$ to the actual number of classes. The performance of the clustering is measured in terms of Normalized Mutual Information (NMI), F-measure (F), Precision (P) and Recall (R). NMI is defined as the ratio of mutual information and the average entropy of clusters and entropy of actual ground truth class labels. F-measure is the harmonic mean of precision and recall. For the retrieval task, we use the Recall@K metric that gives us the percentage of test examples that have at least one K nearest neighbor from the same class. More details on these peformance metrics can be found in \cite{Lifted_structure}.

\paragraph{Comparison to state-of-the-art:} We compare the performance of our method against a few recent deep \textit{unsupervised} feature and metric learning methods from literature. The results of comparisons are shown in Table \ref{vs_deep_all}. In all the datasets, we also report the performance obtained using a random projection matrix used as initialization for our method. As observed, despite starting with a random matrix that leads to severely poor performance, our method significantly improves upon it. Mostly, we outperform the baseline methods.

In Table \ref{CUB_sup}, we also compare our method against a few popular \textit{supervised} deep metric learning approaches in recent literature, on the CUB dataset. We observe that our method performs competitive despite not making use of any class labels.
\begin{table}[h]
\centering
\caption{Comparison against supervised deep methods on CUB}
\label{CUB_sup}
\resizebox{0.6\linewidth}{!}{%
\begin{tabular}{|c|c|ccccc|}
\hline
\textbf{Method} & \textbf{Class Labels} & \textbf{NMI}  & \textbf{R@1}  & \textbf{R@2}  & \textbf{R@4}  & \textbf{R@8}  \\ \hline
Contrastive \cite{contrastive}    & Yes                   & 47.2          & 27.2          & 36.3          & 49.8          & 62.1          \\
DDML \cite{DDML}           & Yes                   & 47.3          & 31.2          & 41.6          & 54.7          & 67.1          \\
Triplet \cite{FaceNet}       & Yes                   & 49.8          & 35.9          & 47.7          & 59.1          & 70.0            \\
Triplet+hard \cite{FaceNet}   & Yes                   & 53.4          & 40.6          & 52.3          & 64.2          & 75.0            \\
Triplet+DAML \cite{Deep_AML}   & Yes                   & 51.3          & 37.6          & 49.3          & 61.3          & 74.4          \\
Triplet+HDML \cite{HDML_CVPR19}   & Yes                   & 55.1          & 43.6          & 55.8          & 67.7          & 78.3          \\
Lifted \cite{Lifted_structure}         & Yes                   & \textbf{56.4} & \textbf{46.9} & \textbf{59.8} & \textbf{71.2} & \textbf{81.5} \\ \hline
\textbf{sRPML(Ours)}            & No                    & 53.1          & 43.7          & 56.5          & 68.3          & 79.4          \\ \hline
\end{tabular}%
}
\end{table}

\begin{figure}[!htb]
\centering
	\includegraphics[width=0.65\linewidth]{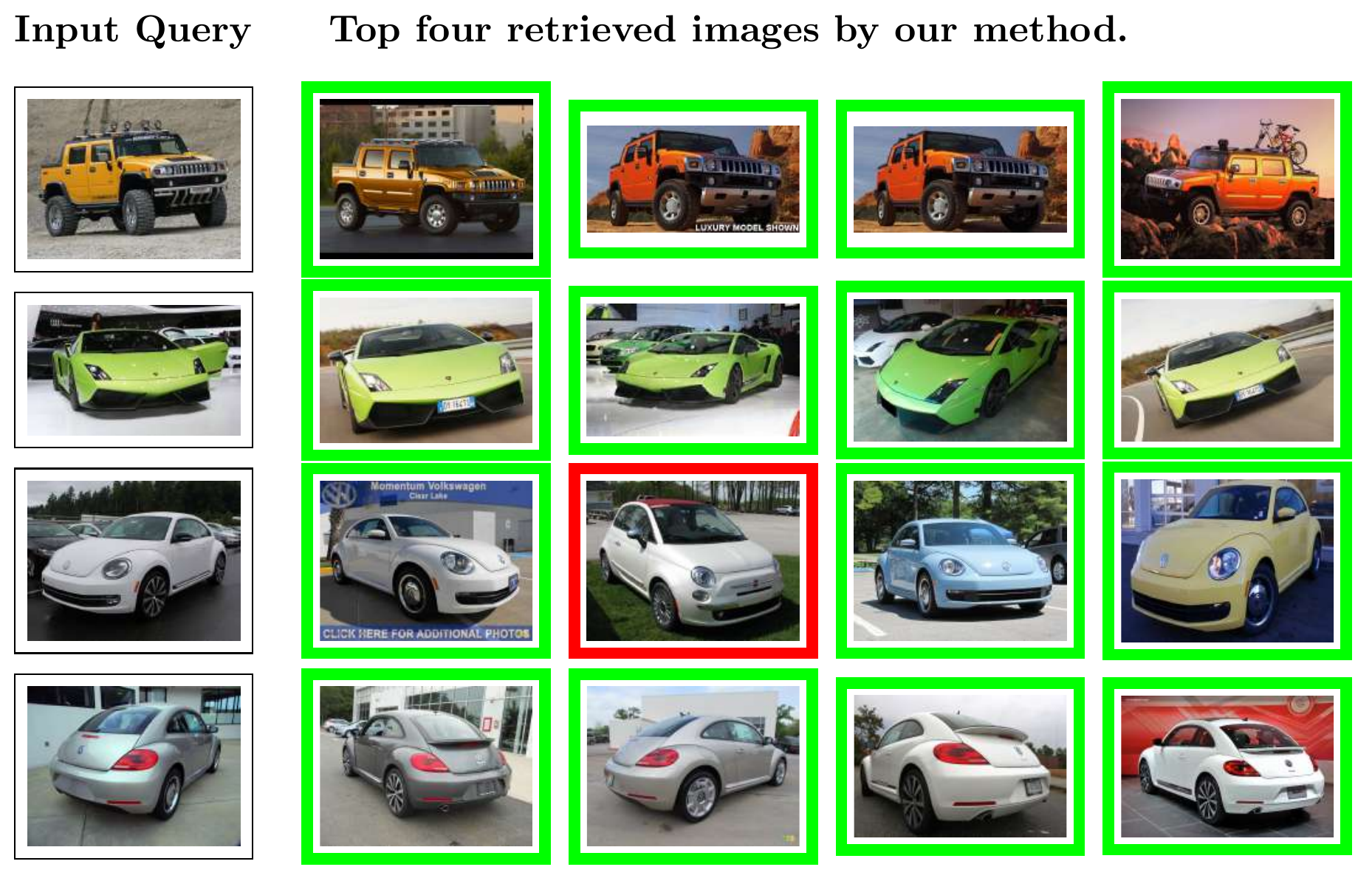}
    \caption{Mostly successful retrieval results for queries from the Cars dataset. The correctly retrieved images are shown in green, while the incorrect ones are shown in red.}
    \label{success_retrieval}
\end{figure}
\begin{figure}[!htb]
\centering
	\includegraphics[width=0.65\linewidth]{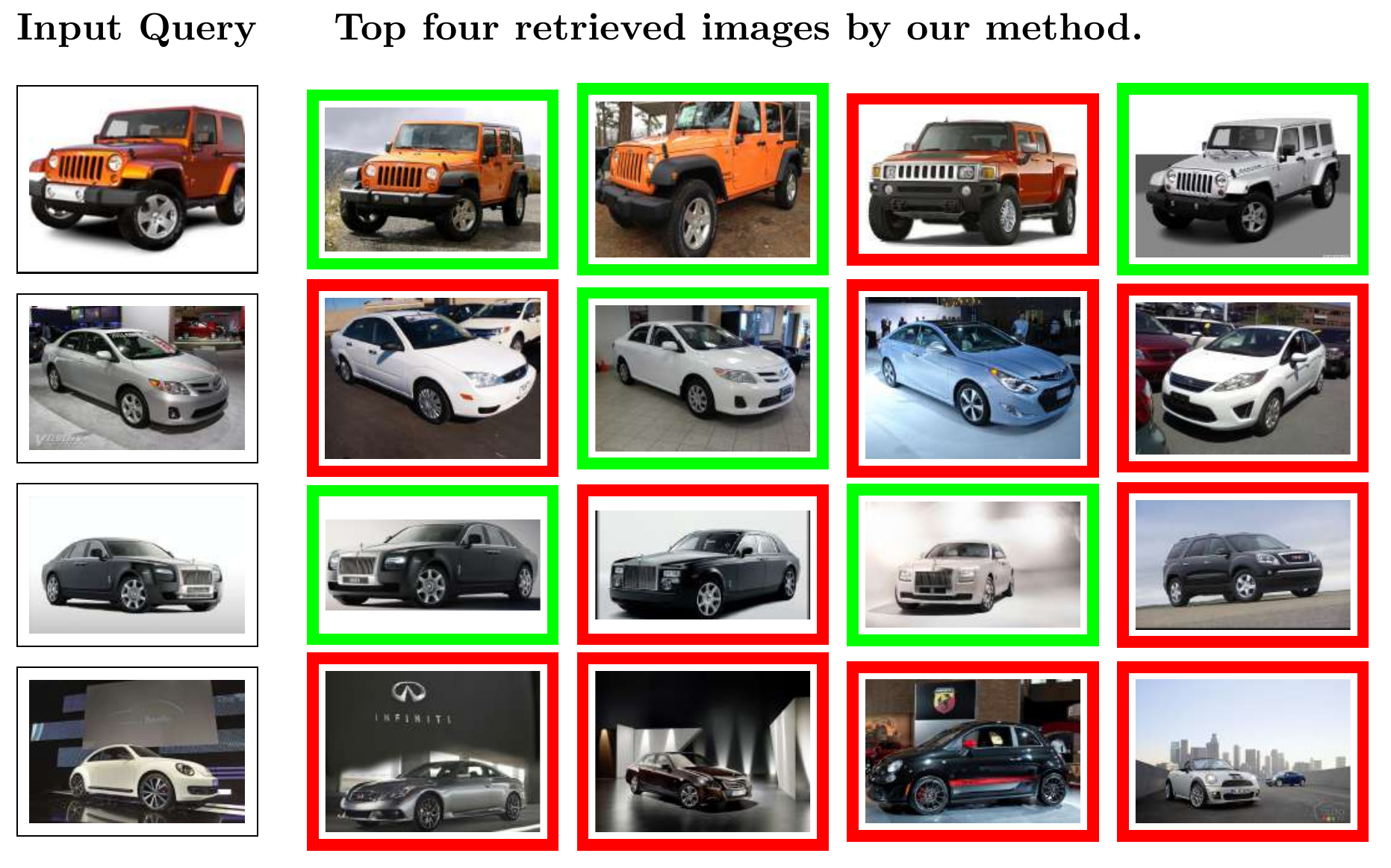}
    \caption{Mostly unsuccessful retrieval results for queries from the Cars dataset. The correctly retrieved images are shown in green, while the incorrect ones are shown in red.}
    \label{failures_retrieval}
\end{figure}

\subsection{Qualitative results on Cars dataset}
\begin{figure}[!htb]
\centering
	\includegraphics[width=0.85\linewidth]{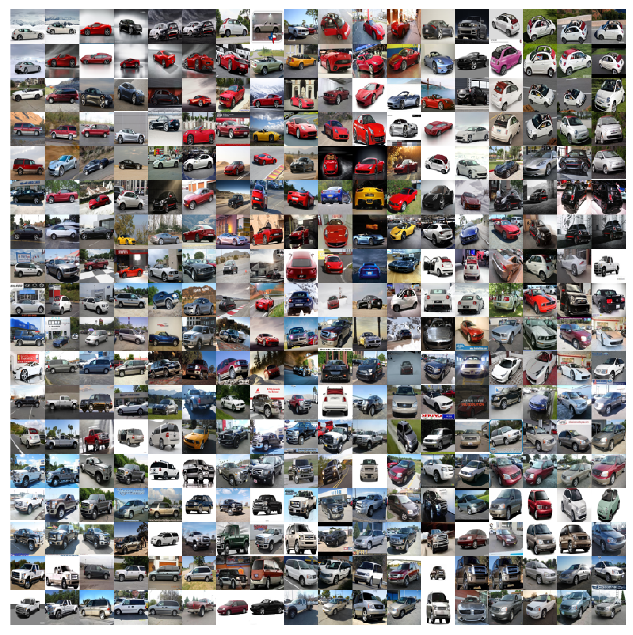}
    \caption{Quad grid layout of Cars196 test images using our method. Similar cars are together. }
    \label{cars_quadgrid}
\end{figure}

We also analyse our proposed method qualitatively on the Cars 196 dataset. For this, we show the retrieval results using our approach in Figures \ref{success_retrieval} and \ref{failures_retrieval}. For each query, we show the top four retrieved images. As observed, the images retrieved are fairly accurate. 
We also show a quad grid layout \footnote{\color{red} https://cs.stanford.edu/people/karpathy/cnnembed/ \color{black}} of nearest neighbors obtained using tSNE embedding of a subset of Cars196 test images, using our method, in Figure \ref{cars_quadgrid}. As observed, similar cars are placed together. To further improve the performance, we may need a few labeled examples to guide the training in a better way. This interesting semi-supervised scenario could be looked at as a future direction of work.

\section{Conclusion}
In this paper we proposed an unsupervised embedding learning approach that uses a graph-based clustering approach to obtain pseudo-labels for examples. The pseudo-labels are used to form triplets of examples, which guide the learning of an embedding. We also propose a weight function to scale the losses associated with these triplets. We jointly learn the parameters of our approach using Riemannian optimization on product manifold, which further ensures a faster convergence. Our approach performs competitive to state-of-the-art metric learning techniques on a number of benchmark datasets. In future, we plan to formulate a semi-supervised variant of our approach to further guide the training process and achieve a better performance.

{
\bibliographystyle{unsrt}
\bibliography{RPML_arXiv19}
}

\end{document}